\documentclass[conference]{IEEEtran}
\IEEEoverridecommandlockouts
\usepackage{cite}
\usepackage{amsmath,amssymb,amsfonts}
\usepackage{graphicx}
\usepackage{textcomp}
\usepackage{xcolor}
\usepackage[font=footnotesize,labelfont=bf]{caption}
\usepackage{hyphenat}
\usepackage{tikz}
\usepackage{pgfplots}
\usepackage{amssymb}
\usepackage{mathtools}
\usepackage{array}
\usepackage{hyperref} 
\usepackage{subcaption}
\usepackage[linesnumbered,ruled,vlined]{algorithm2e}
\usepackage{balance}
\usepackage[utf8]{inputenc}
\usepackage[noend]{algpseudocode}
\usepackage{float} 
\usepackage{colortbl}
\usepackage{algorithmicx}
\hypersetup{
    colorlinks=true,
    linkcolor=blue,
    citecolor=blue,
    urlcolor=blue
}
\usepackage{graphics}
\pgfplotsset{compat=1.18}
\begin{document}
\title{ FedFair\textsuperscript{3}: Unlocking Threefold Fairness in Federated Learning
}
\author
 {Simin Javaherian, Sanjeev Panta, Shelby Williams, Md Sirajul Islam, Li Chen\\School of Computing and Informatics \\
                     University of Louisiana at Lafayette }
\maketitle
\begin{abstract}
Federated Learning (FL) is an emerging paradigm in machine learning without exposing clients' raw data. In practical scenarios with numerous clients, encouraging fair and efficient client participation in federated learning is of utmost importance, which is also challenging given the heterogeneity in data distribution and device properties. 
Existing works have proposed different client-selection methods that consider fairness; however, they fail to select clients with high utilities while simultaneously achieving fair accuracy levels. In this paper, we propose a fair client-selection approach that unlocks threefold fairness in federated learning. In addition to having a fair client-selection strategy, we enforce an equitable number of rounds for client participation and ensure a fair accuracy distribution over the clients. The experimental results demonstrate that FedFair\textsuperscript{3}, in comparison to the state-of-the-art baselines, achieves 18.15\% less accuracy variance on the IID data and 54.78\% on the non-IID data, without decreasing the global accuracy. Furthermore, it shows 24.36\% less wall-clock training time on average. 

\end{abstract}
\section{Introduction}\label{Introduction}

Traditional machine learning relies on centralized servers for data gathering and training, which falls short when dealing with privacy-sensitive data residing in a vast number of edge devices.
To address the challenge, Federated Learning (FL) has emerged as a promising alternative, which operates within a loosely federated network of clients, thereby enabling collaborative model updates. In FL, participants retain their data locally on their devices, perform local training, and subsequently share their model updates with a central server for global aggregation.

Despite its potential for distributed learning over decentralized data, FL has encountered open challenges 
to be addressed for large-scale real-world deployments. One important challenge is to enable a fair system where all clients are encouraged to participate in updating their models. Otherwise, with an unfair system, clients do not receive fair rewards and thus become reluctant to participate.
On the other hand, efficiently learning a high-quality model in FL is challenging, considering the diverse set of client devices and heterogeneous data distributions.
This requires judiciously giving different priorities to clients according to their capabilities of updating models and communicating with servers. 


Recent efforts such as Oort \cite{lai2021oort} have sought to enhance FL performance through guided participant selection, albeit without considering client features and the distribution of accuracy among clients. On the other hand, q-FFL \cite{li2019fair} takes fair accuracy distribution into account but lacks a comprehensive consideration of selective client participation. Focused on communication efficiency, \cite{yu2019parallel, mcmahan2017communication, zhang2012communication, caldas2018expanding, konevcny2016federated} did not consider fairness, inversely impacting client willingness in FL participation. 
All the existing works fail to simultaneously address efficient learning and fair client selection in FL. Moreover, there has not even been any consensus on the definition of fairness.  

To this end, this paper aims to answer the following 
research question: in large-scale FL with heterogeneous clients, how should we elicit and achieve the best possible fairness among all the clients, while ensuring an efficient collaborative model learning? 
This involves addressing the open challenges of defining appropriate fairness constraints and designing an optimal client selection strategy throughout the learning process toward efficiency and fairness goals.  
To address these challenges, we present a novel approach, named \textit{FedFair\textsuperscript{3}}, which takes into account client features while maintaining fairness in client participation from three perspectives, without compromising performance and accuracy. 
In particular, the client features including data size, energy consumption, round duration, power consumption and local loss are comprehensively considered in the probabilistic client selection framework in \textit{FedFair\textsuperscript{3}}. 
Different from the existing probability-based frameworks (\cite{katharopoulos2017biased,goetz2019active}), our approach integrates three simultaneous notions of fairness.  Firstly, participants with similar resources and capabilities are selected with equal probability to ensure fairness in participant selection. Secondly, we introduce the concept of accuracy fairness, where we aim to ensure that the performance of each participant, in terms of the model accuracy level, is proportional to its resource and capability. Thirdly, clients are penalized after a specific number of rounds, to allow fair participation throughout the whole FL process. 

\begin{figure}[!ht]
\centering
\includegraphics[width=5.5cm, height=7.5cm, keepaspectratio]{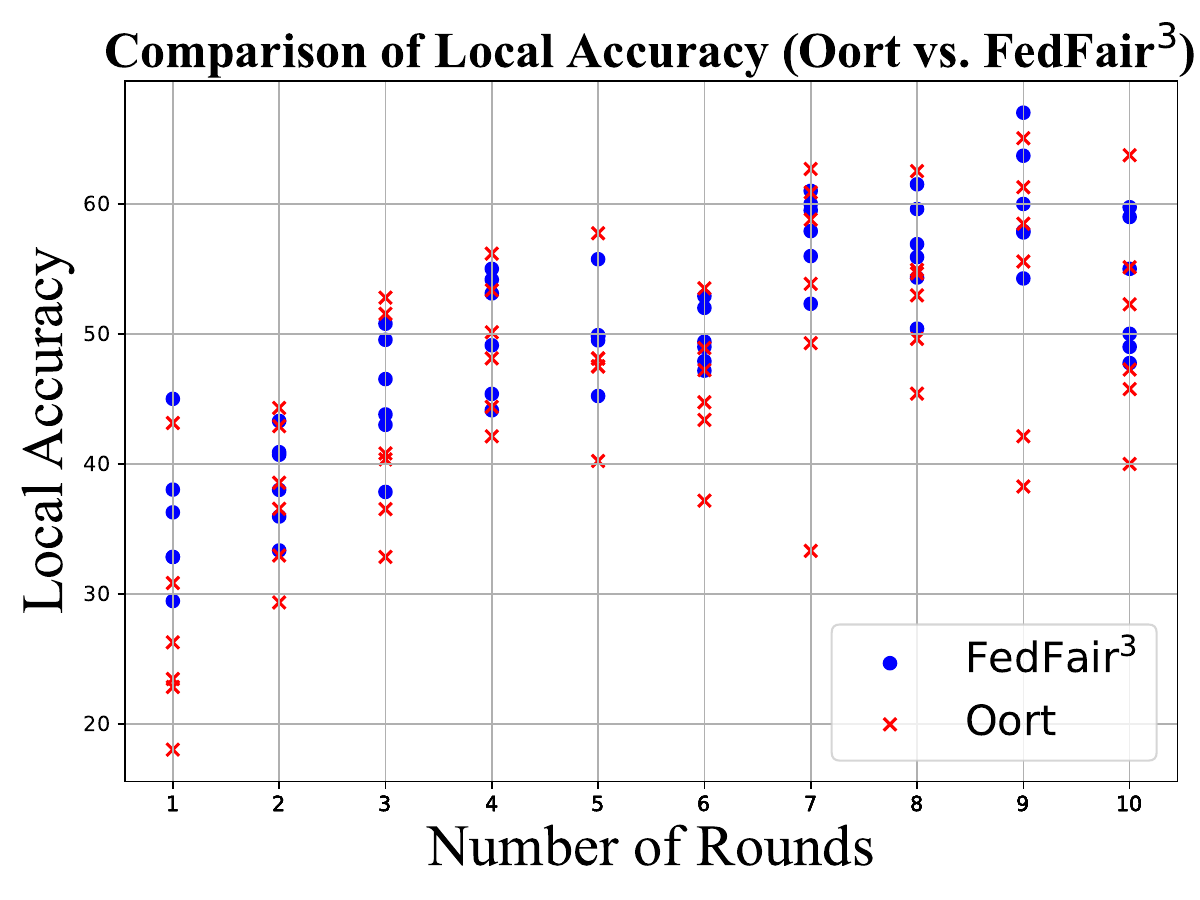}
\caption{A toy example of comparison of local accuracy (Oort vs. FedFair\textsuperscript{3}) considering 6 clients in each round.}
\label{figure1}
\end{figure}

To shed light on this, we provide a glimpse of our results through a toy example on Fig.~\ref{figure1}. Evaluating the accuracy of these clients in each round, it becomes evident that FedFair\textsuperscript{3} significantly outperforms Oort, which serves as the baseline for our approach. The experimental results show that our approach exhibits an 18.15\% reduction in accuracy variance over IID data and a remarkable 54.78\% reduction over non-IID data, showcasing its superiority over existing methods. Furthermore, regarding the efficiency, FedFair\textsuperscript{3} also outperforms the baseline algorithms, achieving a 24.36\% reduction in wall clock time, thereby advancing us closer to the goal of achieving target accuracy in less time, a critical objective in the realm of FL\cite{li2020federated, reddi2020adaptive}.



\textit{Our Contributions are summarized as follows:}
\begin{itemize}
  \item This paper proposes a new approach that selects clients in a fine-grained manner considering their demands and resources including computational power, data size, time duration and energy consumption. 
  \item  It also considers fairness regarding the accuracy variance and the number of rounds that each client can participate in training. We achieve this by giving larger probabilities to those clients that have more capabilities. 
  \item We show the experimental results, demonstrating the deviation of accuracy, the final accuracy and the wall-clock training time for IID and non-IID datasets.
\end{itemize}

\section{Background and Related works}\label{relatedworks}
In this section, we provide an overview of FL, including its general concept and key components. We then summarize and discuss related works to motivate our proposed design. 

\subsection{Preliminaries of Federated Learning } 
FL typically involves a vast amount of edge devices, such as smartphones and laptops, and a server that periodically synchronizes the global model, denoted as $w$, across devices or clients. Considering $f_i(w)$ as the local loss over the $n_i$ data samples held by the client $i$ and $n$ denote the total number of samples, the learning goal in FL is to solve the following optimization problem:

\begin{eqnarray}
 \min_w & & F(w) = \sum \limits_{i=1}^N  {p_i}\alpha_i F_i(w) \label{eq1} \\ 
\text{s.t.} & &
 \nonumber  F_i(w)=\frac{1}{n_i}\sum \limits_{i=1}^{n_i} f_i(w) \\ \nonumber
   & &   \sum \limits_{i} n_i = n, \ \  \nonumber
     \sum \limits_{i} p_i=1, \ \  p_i\geq0 \\ \nonumber
\end{eqnarray}
$p_i$ is the probability of the client selection that usually is $p_i=\frac{n_i}{n}$, and $\alpha_i$ is the weight of client $i$ that is usually considered 1.
The rule of the model update is defined as below:\begin{equation}
w^{t+1}=w^t-\eta \alpha_i \nabla F(w^t) \nonumber
\end{equation}

\subsection{Fairness and Efficiency Related Works}
Fairness in Federated Learning (FL) has been explored from various angles in previous research. The core aim of fairness is to keep all clients engaged and motivated in the learning process. Moreover, using the importance sampling\cite{katharopoulos2017biased}, the probability can be selected as a proportion of loss. That is, $p_i\propto F(w)$. We use the importance sampling in FL, and the experiments show that importance sampling is advantageous over random sampling as it strategically prioritizes training samples based on their loss values, allowing the optimization algorithm to focus more on informative instances, resulting in improved convergence and generalization performance. However, we also take the client utilities and features into account, making a more efficient and fair system. We briefly summarized the different fairness researches as follows.

Counterfactual fairness \cite{kusner2017counterfactual} is another fairness notion in which all individuals, considering the protected attributes in a casual sense, should receive same distribution of prediction, regardless the group they belong to (including race, gender, etc.). Barocas et al. \cite{barocas2017fairness} and Mitchell et al. \cite{mitchell2018prediction} also mentioned that different sensitive groups should receive same patterns of outcomes; otherwise, it would violate the demographic fairness. There are also some surveys which discuss the FL challenges with a focus on fairness issues \cite{shi2021survey,zhan2021survey,zeng2021comprehensive}. 

In optimization, Mohri et al.\cite{mohri2019agnostic} proposed a new framework of agnostic FL that uses a minimax optimization approach, but it optimizes the model just for a small number of clients\cite{li2019fair}. On a larger scale, Li et al. \cite{li2019fair} introduced a parameter q as the q-FFL method that provides a uniform accuracy for the network. AFL is a special case of q-FFL, when the q is large enough. The q-FFL approach minimizes an aggregate reweighted loss parameterized by q such that the devices with higher loss are given higher relative weight. However, it did not consider participant selection. FedProx\cite{li2020federated} incorporates a proximal term into the local training objective with the aim of maintaining the proximity of local models to the global model. 
The authors in \cite{wang2023fedeba+} address the trade-off between fairness and global model performance with using an entropy perspective. Chu et al. in \cite{chu2022focus} propose a formal FL fairness definition, fairness via agent-awareness (FAA), which takes the heterogeneity of different agents into account. In HFFL \cite{zhang2020hierarchically}, agents which contribute more to FL are rewarded more in this framework. The agents at the different contribution levels thus receive different model updates. Fan et al. in \cite{fan2022improving} proposed a new approach that uses a matrix containing clients' contributions and data to address this problem. Another work introduced collaborative fairness in FL (CFFL)\cite{lyu2020collaborative}, which considers fairness regarding the clients' contribution. It evaluates the clients' contributions and updates this information steadily, so it knows the clients' reputation and can fairly distribute updated models. 
Since working with the non-IID data over a distributed system is unfair, Ray et al.\cite{ray2022fairness} introduced a new fairness concept called core-stable fairness. 

\vspace{-1mm}
\subsection{ Motivation}
\textbf{Trade-off between accuracy and fairness.}
The trade-off between accuracy and fairness is an open challenge. We hope to answer the question: is there any solution to improve the fairness without decreasing the accuracy?

\textbf{Considering client features to increase the efficiency and fairness.}
In FL, clients come with varying computational resources, data sizes, and energy limitations, making it crucial to foster fairness. While previous research has introduced various fairness strategies \cite{papadaki2022federated, zhang2020fairfl, ray2022fairness}, our work identifies a gap between fairness and client selection probabilities. 

\textbf{Considering more fairness notions into account.}
Lacking a consensus on fairness, we naturally wonder: Can we have a client selection approach that simultaneously explores more fairness notions?

In response, we propose a novel approach to tackle the multifaceted issue of client selection. Our approach takes a detailed view, considering individual client demands and resources, including computing power, data volume, time constraints, and energy usage. Different from all the existing work, we extend our commitment to fairness beyond just client selection. Our approach also addresses accuracy variance and the number of training rounds allotted to each client, giving preference to those with greater capabilities.


\section{Methodology}\label{Methodology}
In this section, we present our new approach, FedFair\textsuperscript{3}, to address the fairness challenge of client selection in FL. We begin by outlining our assumptions and presenting the objective function,
which serves as the basis for our fairness notions regarding accuracy and participant selection. 
We then present our client selection method, which works in a fair way by using a non-uniform probability distribution instead of a random participant selection. Finally, we show that our approach leads to a uniform and fair system regarding the accuracy variance.

\begin{table}[!ht]
  \centering
    \caption{Notations}
  \renewcommand{\arraystretch}{1.1}
  \resizebox{0.4\textwidth}{!}{
  \begin{tabular}{|l|p{7.3cm}|}
    \hline
    \multicolumn{2}{|c|}{\textbf{The Meaning of the Notations}} \\
    \hline
    \cellcolor{gray!20}T & \cellcolor{gray!20}The developer's preference of duration of each round \\
    \hline
    $t_i$ &  The amount of time that client $i$ takes to perform the training \\
    \hline
   \cellcolor{gray!20}$\tau$ & \cellcolor{gray!20} Local rounds \\
    \hline
    K & The number of global aggregations \\
    \hline
   \cellcolor{gray!20}I & \cellcolor{gray!20}The total number of local iterations \\
    \hline
     $r_i$ &  Duration of local round that client i takes to perform its training \\
    \hline
    \cellcolor{gray!20}$l(\cdot)$ &\cellcolor{gray!20} $\cdot$ is a Boolean value, if it is true, $l(\cdot)$ is 1, otherwise it is 0 \\
    \hline
     $\beta$ &  A penalty value for those clients that take more time to complete their round \\
    \hline
    \cellcolor{gray!20}$d_i$ & \cellcolor{gray!20}The size of the data of the client $i$\\
    \hline
     $c_i$ &  Computational power of the client $i$ \\
    \hline
   \cellcolor{gray!20}$q_i$ & \cellcolor{gray!20}Energy consumption of each client $i$ \\
    \hline
     $\gamma_i$ &  $\in \{0,1\}$, a binary indicator for the selection of client $i$ \\
    \hline
   \cellcolor{gray!20}$\kappa$ & \cellcolor{gray!20}Batch size \\
    \hline
     $s_\vartheta$ &    The current total accumulated type $\vartheta$ resource usage\\
\hline
   \cellcolor{gray!20}$r_l$ & \cellcolor{gray!20} Local resource consumption type $\vartheta$ \\
    \hline
     $r_g$ & Global resource consumption type $\vartheta$ \\
    \hline
   \cellcolor{gray!20}$\Pi_\vartheta$ & \cellcolor{gray!20} Resource budget type $\vartheta$ \\      \hline    L  &  Lipschitz constant \\
  
    \hline 
    \cellcolor{gray!20}E & \cellcolor{gray!20} Explored clients for updating the priority\\
  
    \hline
  \end{tabular}}
  \label{table1}
\end{table}


\textbf{Assumptions:} We assume the following for each client $i$:

\begin{enumerate}

\item $F_i$($w$) is convex\cite{armacki2022personalized}, {\em i.e.}, \begin{equation}
F(\delta w_1 + (1-\delta)w_2) \leq \delta F(w_1) + (1-\delta)F(w_2)
\end{equation}
\item $F_i$($w$) is  L-Lipschitz\cite{wang2019adaptive}, satisfying
 \begin{equation}
     \rVert F_i(w)-F_i(w') \rVert \leq L\rVert w-w'\rVert, \; \forall w, w'
\end{equation}

\end{enumerate}

\vspace{-1mm}

\subsection{FedFair³ Algorithm}
\label{FedDoalgo}
Our approach is designed to ensure fairness and efficiency in the selection of clients for federated learning. We accomplish this by strategically choosing clients with higher losses, while taking into account their distinct features. To achieve a uniform distribution of accuracy across clients, we introduce the concept of weighting in our objective function, denoted as $\alpha_i$. In essence, with this weighting mechanism, each client contributes fairly to the learning process. In more practical terms, our approach not only prioritizes clients with higher loss functions but also factors in their resource capabilities. This means that clients with greater computational resources, larger datasets, lower energy consumption, and quicker training times within our defined limit, denoted as T, have an increased likelihood of being selected. We also ensure that we do not repeatedly select the same clients, thus giving priority to those who have not yet been chosen. 

To achieve this, the server plays a crucial role in aggregating clients' characteristics before initiating the model update. This aggregation allows us to rank clients based on their unique features and available resources, ensuring a well-balanced selection process. However, we also consider time as a valuable resource. Clients that exceed the predefined time limit, $S$, face a penalty denoted as $\beta$, which reduces their selection probability. Conversely, clients that require less time than the set limit have their selection priority determined by their available resources, provided they are selected. In each training round, we carefully monitor the consumption of resources. If the cumulative resource consumption surpasses our predefined budget, the server will signal the clients to stop. This dynamic resource management strategy not only ensures fairness but also maintains the efficiency of the federated learning process. We call the resource budget $\Pi_\vartheta$ for the resource type $\vartheta$. Thus, we have the following resource constraint:
\begin{equation}
(I + 1)r_l + (K + 1)r_g \leq \Pi_\vartheta, \forall \vartheta \;\; s.t.\;\;
I=K\tau    
\end{equation}

Our algorithm relies on the design of client selection probability $p_i$ and client weight $\alpha_i$ for the FL optimization in Eq.~(\ref{eq1}) as follows:
\begin{equation}
S, \;
  \alpha_i = \frac{p_i^q}{N(q+1)}\\
\label{eq7}
\end{equation}
where

\begin{equation}
\begin{multlined}
 \nonumber {U_i}= |\kappa|\sqrt{\frac{1}{|\kappa|}\times  \sum \limits_{i \in \kappa} {Loss(i)^2}}. (\frac{T}{t_i})^{l(T<t_i)\beta}. \lambda^{l(T>t_i)\gamma_i},\\ 
  \lambda= \frac{c_id_i}{q_ir_i},
 \text{$\gamma_i$} =
\begin{cases}
1, & \text{if client i is selected} \\
0, & \text{otherwise}
\end{cases}
 \label{eqgamma}
\end{multlined}
\end{equation}
$U_i$ defines the client \textit{Utility} which refers to a measure that combines the loss values of data samples to assess the clients' significance in improving model performance during various training tasks with $T$ representing the preferred round duration, $t_i$ denoting the time taken by client $i$ to process training data, $\lambda$ as the clients' priority considering the clients features, and $l(x)$ serving as an indicator function that evaluates to $1$ when $x$ is true and $0$ otherwise. Table \ref{table1} demonstrates the meaning of the notations.

We show that with the design of $\alpha_i$ as Eq.~(\ref{eq7}) and substituting in Eq.~(\ref{eq1}), we have the equivalent objective function as in q-fairness \cite{li2019fair}.
\begin{equation}
\begin{multlined}
   \sum \limits_{i=1}^N  {p_i}\alpha_i F_i(w)= \sum \limits_{i=1}^N  {p_i}\frac{p_i^q}{N(q+1)} F_i(w)\\ \propto \sum \limits_{i=1}^N  \frac {p_i F_i(w)^q}{N(q+1)} F_i(w)=\sum \limits_{i=1}^N  \frac {p_i}{N(q+1)} F_i(w)^{q+1} \label{eq2}
   \end{multlined}
\end{equation}

Building upon the concepts of q-fairness, our algorithm aims to address fairness considerations in the context of FL, while also striving for equitable and efficient collaboration among clients. With this context in mind, let's proceed to the summary of Algorithm \ref{FedDo} steps as follows:
\begin{itemize}

\item \textbf{Step1:} The server aggregates clients' features to calculate individual client probabilities for participation in the FL round (Line 3-16). 
\item \textbf{Step2:} Leveraging calculated probabilities, the server prioritizes clients and samples a set based on their priority, penalizing those with lower priority determined by their loss values (Line 17).
\item \textbf{Step3:} Models are sent to the sampled clients, emphasizing those with higher priority, enabling them to contribute to the federated model update (Line 18-22).
\item \textbf{Step4:} The server checks the resource budget by considering previous values and current available resources, ensuring that the federated learning process aligns with resource constraints (Line 23-24).
\item \textbf{Step5:} 
Clients, whose participation adheres to the resource budget, engage in model training, promoting collaboration while respecting resource limitations (Line 29-32).
\end{itemize} 

\SetKwInput{KwInit}{Initialization}
\SetKwInput{KwOutput}{Output}  
\begin{algorithm}[!ht]
\small
\DontPrintSemicolon
\SetAlgoNlRelativeSize{-2.5}
\SetAlgoLined
\textbf{Participant selection for each round}
\hrule
\caption{FedFair\textsuperscript{3}}\label{FedDo}

\textbf{At The Aggregator}\\
\While{True}{
  Alleviating preferred time T\\
   Calculate client probability: \\ 
  \For{client $i \in E$}{
    calculate $Loss F(i)$\;
    $U_i \gets |\kappa| \sqrt{\frac{\sum_{i \in \kappa}F(i)^2}{|\kappa|}}$\;
    \eIf{$T < t_i$}{
      $U_i \gets U_i \times \left(\frac{T}{t_i}\right)^\beta$\;
    }
    {
      $\lambda \gets \frac{c_id_i}{q_ir_i}$\;
      $U_i \gets U_i \times \lambda$\;
    }
    $p_i=\frac{U_i}{\sum U_i} $ \\
  }
 sample clients by priority\\
   \For{selected clients}{
    Receive $p_i, \alpha_i, \nabla F_i(w), c_\vartheta$ for all $i$\;
    $\nabla F(w) \gets \nabla F(w) + p_i\alpha_i\nabla F_i(w)$\;
    $w^{t+1} \gets w^t - \eta \alpha_i \nabla F(w^t)$\;
  }
  \If{$\exists \vartheta | s_\vartheta + \sum_{j \in N} {(r_{j,\vartheta}\tau + 2r_g)} > \Pi_\vartheta$}{
    Send a "Stop" message to the clients\;
  }
  Return $w$\;
}

\textbf{At the selected clients}\\
\While{they did not receive a "Stop" message from the aggregator}{
  Receive $w$ and $\tau$\;
  Perform updating $w_i$\;
  Send $w_i$, $\nabla F(w_i)$, $r_{\vartheta,i}$\;
  }
\end{algorithm}

\subsection{Fairness Qualification}\label{qualification}
This section proves that our objective function has a uniform distribution over clients. For the ease of mathematical exposition, we consider the following objective function of FL with limiting q to 0 as a conventional objective function or 1 as our objective function: $ min_w F_q(w)=\sum \limits_{i=1}^N {p_i\alpha^q_i}F_{i,q}(w)^{q+1}$.

We show this uniform distribution with two definitions as follows:
 
\textbf{Definition 1}. \textit{Uniformity of variance of the performance distribution:} we say the distribution of $N$ clients $\{F_{1,q}(w),..., F_{N,q}(w)\}$ under solution $w$ is more uniform than $w'$ if:
\begin{equation}
\begin{multlined}
\mathbf{Var_p}[ F_{1,q}{(w)},..., F_{N,q}{(w)}]< \mathbf{Var_p}[ F_{1,q}{(w')},..., F_{N,q}{(w')}]
\end{multlined}
\end{equation}

Our algorithm selects clients with higher losses. As a result, we have a system with less variance.
According to our first notation of fairness, our system is fair regarding its variance.

\textbf{Definition 2}. \textit{Uniformity of cosine similarity between the performance distribution and \textbf{1}:} we say the distribution of cosine similarity over $N$ clients $\{F_{1,q}(w),..., F_{N,q}(w)\}$ under solution $w$ and \textbf{1} is more uniform than $w'$ and \textbf{1} if: 

\begin{equation}
\begin{multlined}
\frac{\frac{1}{N}\sum \limits_{i=1}^N F_{i,q}(w)}{\sqrt{\frac{1}{N}\sum \limits_{i=1}^N F^2_{i,q}(w)}} \geq   \frac{\frac{1}{N}\sum \limits_{i=1}^N F_{i,q}(w')}{\sqrt{\frac{1}{N}\sum \limits_{i=1}^N F^2_{i,q}(w')}}
\end{multlined}
\end{equation}
Considering $w^*$ as the optimal solution of $min \, \, F_q(w)$, we have: $\frac{1}{N}\sum \limits_{i=1}^N F_{i,1}(w^*) \geq   \frac{1}{N}\sum \limits_{i=1}^N F_{i,0}(w^*)$ and $\frac{1}{N}\sum \limits_{i=1}^N F_{i,1}(w^*)^2 \geq   \frac{1}{N}\sum \limits_{i=1}^N F_{i,0}(w^*)^2$. Omitting similar steps as in \cite{li2019fair} due to the space limit, we have: \begin{equation}\nonumber \frac{\frac{1}{N}\sum \limits_{i=1}^N F_{i,1}(w^*)}{\sqrt{\frac{1}{N}\sum \limits_{i=1}^N F^2_{i,1}(w^*)}} \geq   \frac{\frac{1}{N}\sum \limits_{i=1}^N F_{i,0}(w^*)}{\sqrt{\frac{1}{N}\sum \limits_{i=1}^N F^2_{i,0}(w^*)}}\end{equation}

\begin{figure*}[!htbp]
\centering
{\includegraphics[height=3.1cm, width=0.24\textwidth]{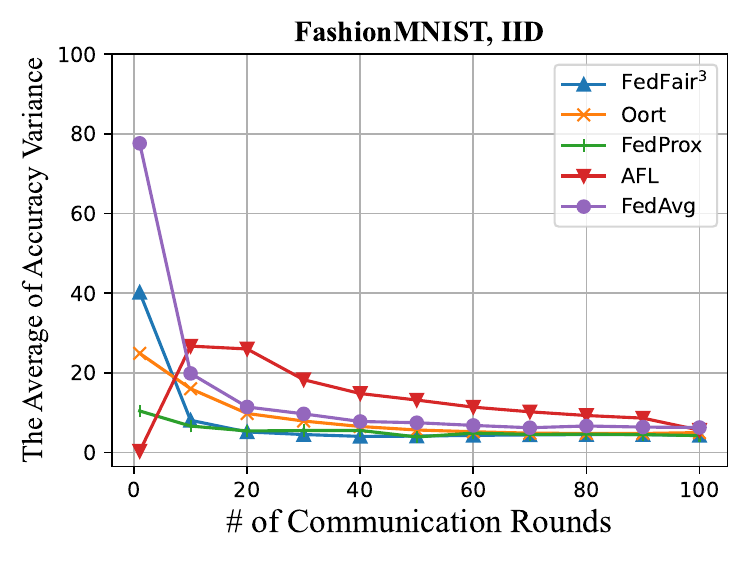}}
{\includegraphics[height=3.1cm, width=0.24\textwidth]{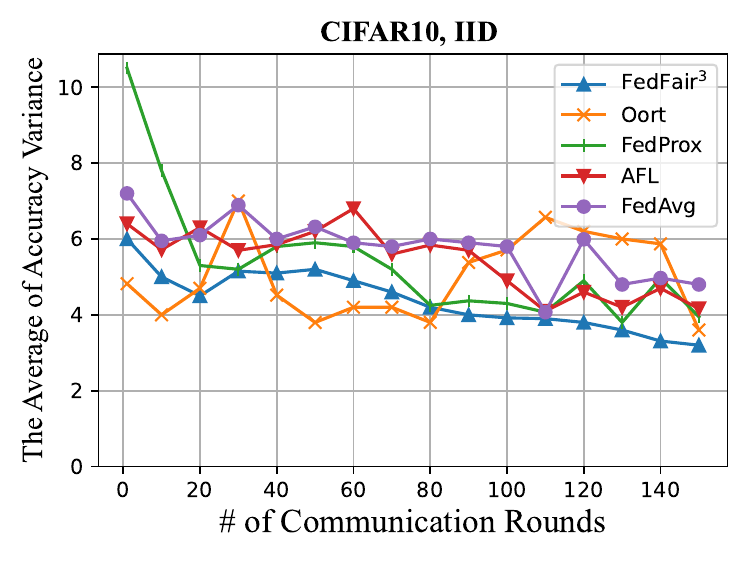}}
{\includegraphics[height=3.1cm, width=0.24\textwidth]{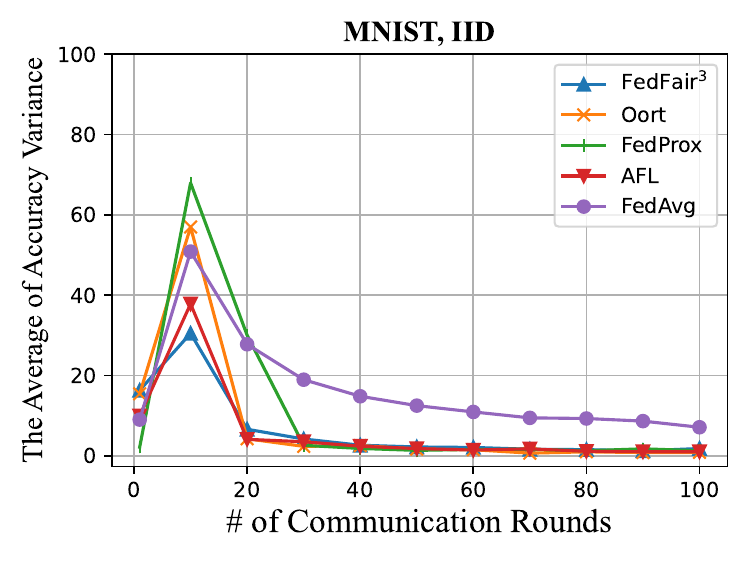}}
\\
\centering
{\includegraphics[height=3.1cm, width=0.24\textwidth]{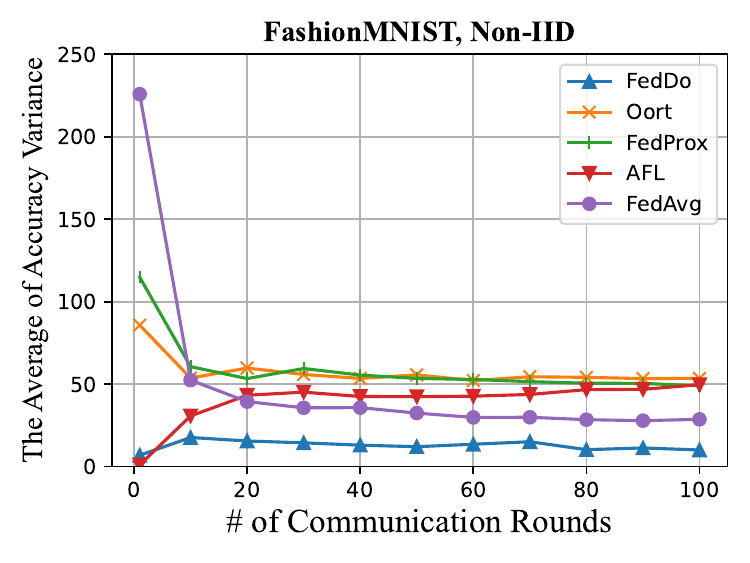}}
{\includegraphics[height=3.1cm, width=0.24\textwidth]{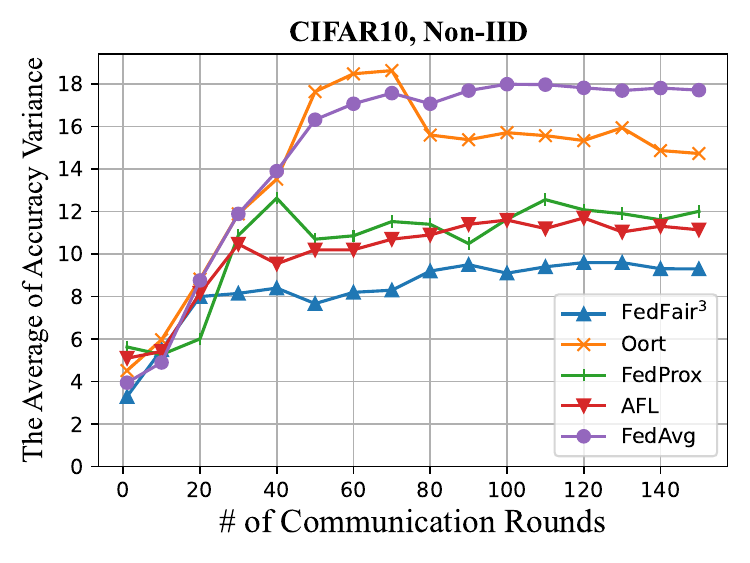}}
{\includegraphics[height=3.1cm, width=0.24\textwidth]{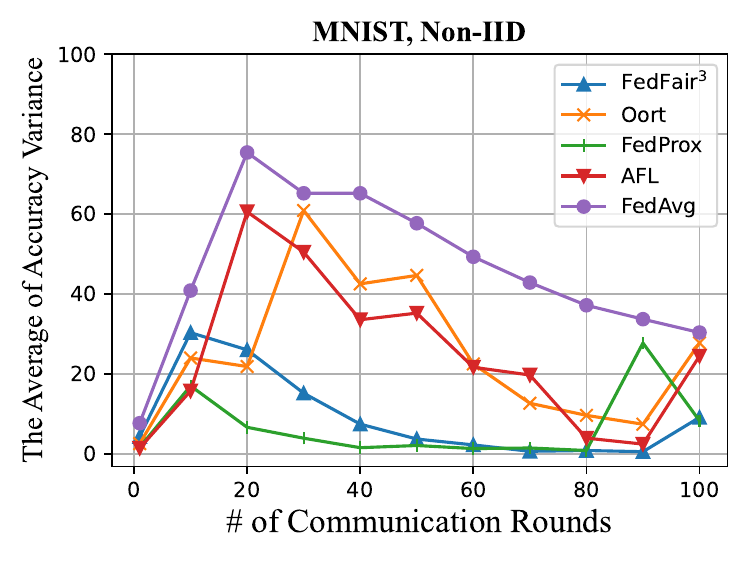}}
\caption{The variance of accuracy of FedFair\textsuperscript{3} versus the FedAvg, AFL, FedProx and Oort.}
\label{figure4}
\end{figure*}

\section{Experimental Result}\label{experimental}
\subsection{Experimental setup}
\textbf{Dataset and Models.} We evalute the performance of FedFair\textsuperscript{3} on three different popular benchmark datasets including MNIST \cite{deng2012mnist}, FashionMNIST \cite{xiao2017fashion}, and CIFAR10 \cite{krizhevsky2009learning} on IID and non-IID datasets.

\textbf{Implementation}.
 Our experiments were conducted using the Plato framework \cite{plato, suasynchronous}. We ran each experiment three times using the table \ref{table3} parameters with a penalty factor 2.
 We executed all experiments on a server, which is equipped with NVIDIA GeForce RTX 3080Ti GPU, Intel(R) Core(TM) i9-10900X CPU, and 64G RAM. We run both the server and clients on the same machine, a configuration supported by the fact that the performance metrics we evaluate are independent of the physical separation between the server and clients.

\begin{table}[!htbp]
\caption{The objective used for experimental results}
\vspace{-3mm}
\begin{center}
\renewcommand{\arraystretch}{1.0} 
\resizebox{0.38\textwidth}{!}{
\begin{tabular}{|c|c|c|c|}\hline
\multicolumn{4}{|c|}{\textbf{Hyperparameters}}\\\hline      
\cellcolor{gray!20}Objectives&\cellcolor{gray!20} MNIST & \cellcolor{gray!20}CIFAR10 & \cellcolor{gray!20}FashionMNIST\\\hline     
     q & 2 & 2 &  2 \\
      \cellcolor{gray!20}Learning Rate & \cellcolor{gray!20}0.01& \cellcolor{gray!20}0.001 &  \cellcolor{gray!20}0.01 \\    
      \# Clients/Round & 10  & 20 & 6 \\     
      \cellcolor{gray!20}\# Total Clients &\cellcolor{gray!20}100 &\cellcolor{gray!20} 300  & \cellcolor{gray!20}100\\     
      \# Rounds & 100  & 150 & 100 \\     
      \cellcolor{gray!20}Batch Size &\cellcolor{gray!20}100 &\cellcolor{gray!20} 64  &\cellcolor{gray!20} 100\\ 
        Optimizer &SGD & Adam  & Adam\\     
      \cellcolor{gray!20}Model Name & \cellcolor{gray!20}LeNet-5 \cite{lecun1989backpropagation}  &\cellcolor{gray!20} VGG16 \cite{simonyan2014very} & \cellcolor{gray!20}LeNet-5 \cite{lecun1989backpropagation} \\  \hline  
\end{tabular}
}
\label{table3}
\end{center}
\caption{The wall clock time of FedFair$^3$ vs. baseline algorithms.}
\vspace{-3mm}
\begin{center}
\renewcommand{\arraystretch}{1.0} 
\resizebox{0.38\textwidth}{!}{
\begin{tabular}{|c|c|c|c|c|}\hline
\multicolumn{5}{|c|}{\textbf{Wall Clock Time Comparison(in second)}}\\\hline  \cellcolor{gray!20} Dataset & \cellcolor{gray!20} AFL &  \cellcolor{gray!20}FedProx & \cellcolor{gray!20}Oort & \cellcolor{gray!20}FedFair\textsuperscript{3} \\\hline    
      FMNIST IID & 16908&  16686 & 16683&\textbf{16568}\\
       \cellcolor{gray!20}FMNIST Non-IID& \cellcolor{gray!20}33598 & \cellcolor{gray!20}33073  &  \cellcolor{gray!20}33307  &  \cellcolor{gray!20}\textbf{33070} \\    
       MNIST IID  & 30458 & 2317& 2932&\textbf{2198}  \\     
     \cellcolor{gray!20} MNIST Non-IID &\cellcolor{gray!20}15954 &\cellcolor{gray!20} 3818  & \cellcolor{gray!20}2332& \cellcolor{gray!20}\textbf{2281}\\     
      CIFAR10 IID & 12199  & 12414 & 12701&\textbf{12040} \\     
      \cellcolor{gray!20}CIFAR10 Non-IID&\cellcolor{gray!20}12712 &\cellcolor{gray!20} 16276  &\cellcolor{gray!20} 12404&\cellcolor{gray!20} \textbf{12068}\\      
  \hline  
\end{tabular}}
\label{table5}
\end{center}
\end{table}

\subsection{Results and Analysis}
\textbf{Performance Comparison.}
Table \ref{table4} and Fig.~\ref{figure4}, demonstrate the results of our experimental evaluation. We compared the proposed approach, FedFair\textsuperscript{3}, with the baseline algorithms: Oort, FedProx, AFL and FedAvg. As depicted, the observed results indicate that there is not a substantial difference in terms of accuracy between the compared methods. However, it is worth noting that our algorithm demonstrates a notable advantage in terms of reduced variance of accuracy across different clients. Furthermore, based on the data provided in the table \ref{table4}, it can be observed that the proposed algorithm, on the average, exhibits 21.31\% less accuracy variation in CIFAR10, 48.48\% less variation in FashionMNIST, and 45.11\% less variation in MNIST non-IID datasets compared to the Oort algorithm.
In addition to the improved accuracy variance, our proposed algorithm, FedFair\textsuperscript{3}, also offers enhanced efficiency by considering the features of clients during the client selection process. As illustrated in table  \ref{table5}, it can be observed that FedFair\textsuperscript{3} exhibits lower wall clock time, on the average 24.36 \%, compared to other algorithms, particularly in non-IID data scenarios. This indicates that FedFair\textsuperscript{3} achieves faster execution and demonstrates its advantage in terms of time efficiency.


\begin{table}[!htbp]
\caption{Global accuracy and accuracy variance for IID and non-IID datasets across MNIST, CIFAR10, and FashionMNIST}
\vspace{-3mm}
 \footnotesize
\begin{center}
\renewcommand{\arraystretch}{0.95} 
\resizebox{0.45\textwidth}{!}{
\begin{tabular}{c|c|c|c|c}\hline
   \multicolumn{5}{c}{\textbf{CIFAR10 Dataset}}\\\hline
   
   \multicolumn{3}{c|}{\textbf{IID}} & \multicolumn{2}{c}{\textbf{Non-IID}} \\\hline

      Alg. & Accuracy & Variance & Accuracy & Variance \\\cline{1-5}
      FedAvg & 86.82 $\pm$0.7 & 4.81 $\pm$0.61 & 76.74 $\pm$0.87& 17.71$\pm$1.02\\
      AFL &  86.91 $\pm$0.98& 4.16$\pm$0.4 & 76.1$\pm$0.3 & 11.01$\pm$1.1\\
      FedProx & 86.22 $\pm$0.34& 3.95 $\pm$0.92& 76.25$\pm$0.88& 14.73$\pm$1.15 \\
      Oort & 86.13 $\pm$0.97 & 3.59 $\pm$0.83 & 75.16 $\pm$0.07 & 13.13 $\pm$0.54\\
      FedFair\textsuperscript{3} & \textbf{86.93} $\pm$0.88& \textbf{3.25 }$\pm$0.6  & \textbf{76.98}$\pm$0.38 & \textbf{ 10.32} $\pm$0.3 
\end{tabular}}
\resizebox{0.45\textwidth}{!}{
\begin{tabular}{c|c|c|c|c}\hline
   \multicolumn{5}{c}  {\textbf{FashionMNIST Dataset}} \\\hline
     \multicolumn{3}{c|}{\textbf{IID}} & \multicolumn{2}{c}{\textbf{Non-IID}} \\\hline

      Alg. & Accuracy & Variance & Accuracy & Variance \\\cline{1-5}
        FedAvg &  78.25$\pm$0.35 & 6.27$\pm$1.03 & 76.31$\pm$0.75 & 27.7$\pm$0.74\\
      AFL & 79.58$\pm$0.75 & 5.62 $\pm$0.93& 74.11$\pm$0.93 & 56.47$\pm$1.35\\
      FedProx & 79.72 $\pm$0.2& 4.23 $\pm$0.44 &76.5$\pm$0.93 &53.77$\pm$1.03 \\
      Oort & 75.95$\pm$0.67 & 5.01 $\pm$0.65 & 75.73 $\pm$0.98& 54.98$\pm$1.2\\
      FedFair\textsuperscript{3} & \textbf{80.79} $\pm$0.84& \textbf{4.25}$\pm$0.74  & \textbf{76.8}$\pm$0.99& \textbf{10.13}$\pm$0.93
\end{tabular}}

\resizebox{0.45\textwidth}{!}{
\begin{tabular}{c|c|c|c|c}\hline
   \multicolumn{5}{c}{\textbf{MNIST Dataset}}\\\hline
   
   \multicolumn{3}{c|}{\textbf{IID}} & \multicolumn{2}{c}{\textbf{Non-IID}} \\\hline

      Alg. & Accuracy & Variance & Accuracy & Variance \\\cline{1-5}
      FedAvg &  96.69 $\pm$0.95& 6.79$\pm$1.07 & 89.27$\pm$0.92 & 27.61$\pm$1.24\\
      AFL &  96.59$\pm$0.89 & 5.69 $\pm$1.39& 88.92$\pm$0.98 & 24.46$\pm$1.04\\
      FedProx & 96.23 $\pm$0.85& 8.27$\pm$1.03& 85.01 $\pm$0.76& 10.27$\pm$0.95 \\
      Oort & 96.46 $\pm$1.18 & 7.56 $\pm$0.93 & 86.73 $\pm$0.83 & 27.78 $\pm$1.3\\
      FedFair\textsuperscript{3} & \textbf{96.74 }$\pm$0.8& \textbf{5.15 }$\pm$1.23  & \textbf{89.79}$\pm$0.78 & \textbf{ 9.02} $\pm$1.52 \\\hline
\end{tabular}
\label{table4}}
\end{center}
\end{table}
\section{Limitation and Future Work}\label{discussion} 
In this section, we outline the limitations of our approach. One of the assumptions of our paper is that $F(w)$ is a convex function, which can be limiting in real-world scenarios. So, we plan to enhance its practical applicability by handling non-convex functions. 
Furthermore, because of the simplicity of the FedAvg, we excluded this algorithm from the elapsed time figures. The wall clock time of the FedFair\textsuperscript{3} is slightly more than FedAvg algorithm in some cases. Another challenge of the approach was finding an optimal value for hyper parameters such as preferred T or q; we used some specific values of q to check the result. It is more efficient to check the q value adaptively. The other limitation of our work is computation overhead. We will try to achieve same results with less computations in the future.

\section{Conclusion}\label{conclusion}
In conclusion, our study has highlighted the importance of fairness in FL and proposed a novel approach to achieve the goal by incorporating client resources and demands. By introducing a weighted loss in the FedFair\textsuperscript{3} algorithm, we have ensured a fair distribution of accuracy over the clients while optimizing resource utilization and improving system performance. Our experimental results have demonstrated that the FedFair\textsuperscript{3} algorithm outperforms existing methods including AFL, FedProx, Oort and FedAvg by achieving a more uniform distribution of accuracy over the clients. 
By considering client resources and demands, the FedFair\textsuperscript{3} algorithm has been successful in achieving fairness and efficiency in distributed machine learning.


\balance

\bibliographystyle{IEEEtran}
\bibliography{Bib}

\end{document}